\documentclass[journal]{IEEEtran}
\usepackage[english]{babel}
\usepackage[T1]{fontenc}
\usepackage[utf8]{inputenc}
\usepackage[ruled, vlined,linesnumbered]{algorithm2e}
\usepackage{authblk}
\usepackage[numbers,sort&compress]{natbib}
\usepackage{pdflscape}
\usepackage[space]{grffile}
\usepackage[cmex10]{amsmath}
\usepackage{amsthm}
\usepackage{amssymb}
\usepackage{xspace}
\theoremstyle{definition}
\newtheorem{definition}{Definition}[section]
\usepackage{multirow}
\usepackage{graphicx}
\usepackage{booktabs}
\usepackage[inline]{enumitem}
\usepackage{tabularx}
\providecommand{\orcidID}[1]{}
\usepackage{todonotes}
\usepackage{comment}
\usepackage{algorithmic}

\usepackage{etoolbox} % \BeforeBeginEnvironment
\BeforeBeginEnvironment{appendices}{\clearpage}
\PassOptionsToPackage{hyphens}{url}
\usepackage{url}
\usepackage[allcolors=blue,bookmarks,colorlinks]{hyperref}

\usepackage{newrevisor}
\newrevisor{manuel}{Purple}
\newrevisor{mahdi}{OliveGreen}
\newrevisor{richard}{Blue}

\renewcommand{\vec}[1]{\ensuremath{\mathbf{#1}}}

\newcommand{\abs}[1]{\ensuremath{\lvert#1\rvert}}

\newcommand{\Fdm}{\ensuremath{F_\textsc{dm}}}

\newcommand{\Ninteractions}{\ensuremath{N_\textup{int}}}
\newcommand{\Nexa}{\ensuremath{N_\text{exa}}}

\usepackage{array}
\newcolumntype{H}{>{\setbox0=\hbox\bgroup}c<{\egroup}@{}}
\usepackage{orcidlink}

\begin{document}
\bstctlcite{BSTcontrol}
\title{Dynamic Detection of Relevant Objectives and Adaptation to Preference Drifts in Interactive Evolutionary Multi-Objective Optimization}
% \author[]{    }
% \author{Blind Review}
\author{Seyed Mahdi Shavarani\orcidlink{0000-0002-3316-1252}}
\author[2]{Mahmoud~Golabi\orcidlink{0000-0002-3314-2051},~\IEEEmembership{Member,~IEEE}}
\author[2]{Lhassane Idoumghar\orcidlink{0000-0001-8853-3968}}

\author[3]{Richard~Allmendinger\orcidlink{0000-0003-1236-3143},~\IEEEmembership{Senior Member,~IEEE}}
\affil{Centre~for~Logistics~\&~Sustainability Analytics,~Kent~Business~School,~University~of~Kent, Kent, UK \authorcr Email: m.shavarani@kent.ac.uk}
\affil[2]{Université~de~Haute-Alsace,~IRIMAS~UR~7499, F-68100~Mulhouse,~France}
\affil[3]{Alliance~Manchester~Business~School,~University~of~Manchester, Manchester,~M13~9PL,~UK}

% \author{Richard Allmendinger}
% \orcid{0000-0003-1236-3143}

\maketitle

\begin{abstract}
%NEW SHORTENED VERSION 

Evolutionary Multi-Objective Optimization Algorithms (EMOAs) are widely employed to tackle problems with multiple conflicting objectives. Recent research indicates that not all objectives are equally important to the decision-maker (DM). In the context of interactive EMOAs, preference information elicited from the DM during the optimization process can be leveraged to identify and discard irrelevant objectives, a crucial step when objective evaluations are computationally expensive. However, much of the existing literature fails to account for the dynamic nature of DM preferences, which can evolve throughout the decision-making process and affect the relevance of objectives. This study addresses this limitation by simulating dynamic shifts in DM preferences within a ranking-based interactive algorithm. Additionally, we propose methods to discard outdated or conflicting preferences when such shifts occur. Building on prior research, we also introduce a mechanism to safeguard relevant objectives that may become trapped in local or global optima due to the diminished correlation with the DM-provided rankings. Our experimental results demonstrate that the proposed methods effectively manage evolving preferences and significantly enhance the quality and desirability of the solutions produced by the algorithm. %This work advances interactive multi-objective optimization by offering a framework that more accurately reflects evolving DM preferences, setting a new standard for adaptive optimization in complex decision-making scenarios.

\end{abstract}


\section{Introduction}\label{introduction}
In many real-world optimization problems, candidate solutions often involve numerous numerical features that need simultaneous optimization. Each of these features holds varying degrees of importance for decision makers, making it challenging to prioritize objectives. This often leads to the inclination to model as many objectives as possible~\cite{SinSaxDeb2013asc}. However, such an approach significantly increases computational time, which increases exponentially with the number of objectives~\citep{Jen03,AllJasLieTam2022cor, BeuFonLopPaqVah09:tec}.

On the other hand, multi-objective problems are characterized by conflicting objective functions, aiming to identify a set of solutions that offer compelling trade-offs, termed non-dominated or Pareto-optimal solutions. The count of Pareto-optimal solutions grows exponentially with the number of objectives~\cite{PurFle2003cec,SinIsaTap2011pareto}, further complicating the decision-making process~\cite{BroZit2009ec}. Thus, it is essential to reduce the number of objectives, if possible.

Previous studies on objective reduction have focused mainly on eliminating objectives that are highly correlated with others~\cite{BroZit2009ec, SinSaxDeb2013asc} or those that do not significantly impact the dominance relations among solutions~\cite{brozit2006dimensionality, BroZit2006allobjectives}. However, irrespective of the problem structure, certain objectives may be ``irrelevant'' to the decision maker (DM) and can be excluded from the optimization model. In addition, there are instances where the DM preferences are influenced by factors beyond the optimized objectives, including numerical features observed by the DM but not explicitly targeted by the system. These features, which are ``hidden'' from the optimizer but relevant to the DM, can affect the satisfaction of the DM if neglected during optimization~\cite{BatCam2009reactive, KryMulPen2022hidden}. The concept of hidden objectives was previously discussed under the term ``unmodeled criteria'' by~\citet{Ste1996jmcda} and was defined as objectives that exist within the internal utility function (UF) of the DM, but are not included in the preference model. Later, formal definitions of hidden and irrelevant objectives were proposed in~\cite{shavarani2023detecting}.

The detection of hidden and irrelevant objectives is possible using the preference information obtained in interactive evolutionary multi-objective optimization algorithms (iEMOA)~\citep{JasBran2008}. iEMOAs are designed to address the significant performance decline in traditional EMOAs caused by the increase in the number of objectives, which reduces the selection pressure~\cite{khare2003performance, ishibuchi2008evolutionary, deb2005scalable}. iEMOAs overcome this issue by utilizing the DM's preferences and discrimination information to develop a preference model. This preference model is used mainly to break the ties between solutions in the same rank and to generate only the parts of the Pareto front (PF) that are interesting to the DM~\cite{branke2008multiobjective, greco2010interactive}. By alternating between decision-making and optimization phases and exploiting the DM's preference information, iEMOAs minimize computational costs and support the DM in finding a desirable solution with minimal cognitive effort. 

It's common for a decision maker (DM) to adjust preferences during the optimization process, often as a result of learning and exploring the solution space through interactions~\cite{sinha2011progressively, thiele2009preference, CamPas2010lion}. This phenomenon, known as "preference drift"\cite{PuChen2008}, affects the significance and relevance of objective functions, highlighting the need to account for the dynamic nature of DM preferences\cite{taylor2018interactive, 985691}. 
Objective sets can also change due to external factors. This situation arises in dynamic multi-objective optimization problems (DMOPs), where constraints, objectives, and parameters evolve over time, potentially altering the problem's Pareto front and even the number of objectives~\cite{farina2004dynamic, ruan2024knowledge}. This dynamic aspect is evident in various real-world scenarios, such as balancing project costs with minimizing the makespan in project scheduling as deadlines approach, incorporating the objective of reducing energy consumption when relying on battery power in supply systems, or incorporating flood-related cost reduction into water resource management systems that were initially designed to minimize construction costs, in response to the impacts of climate change \cite{ruan2024knowledge,CheDinYan2020dynamic}.

% \textcolor{red}{In the context of iEMOAs, the DM's region of interest can be viewed as a discrete dynamic parameter that can shift during optimization. Although this does not change the fundamental nature of the problem or its Pareto front, it necessitates a response to the evolving requirements. However, in an interactive environment where the DM can modify the algorithm’s focus, significant preference changes may pose challenges. Focusing the search on a specific Pareto front region can lead to a lack of solution diversity. Substantial preference shifts might slow the search or cause it to get 'stuck' in a local optimum of a multi-modal problem \cite{8789949}.}

For the first time,~\citet{shavarani2023detecting} suggested using the valuable preference information not only for the direction of the search, but also for dynamic refinement of the set of objective functions. However, they did not address the dynamic nature of the DM's preferences, which significantly contributes to the emergence of both irrelevant and hidden objectives, as these preferences evolve through interactions~\cite{CamPas2010lion}.  

Based on~\cite{shavarani2023detecting}, which served as a pilot study to detect hidden and irrelevant objectives in iEMOAs, the present work makes several significant contributions to the field considering the dynamic preference behavior of DMs. First, this study simulates various levels of preference drift and analyzes their effects on the performance of the detection technique. Second, it discusses a methodology aimed at mitigating the effects of preference changes when they occur. Third, a new method is proposed to detect preferences changes based on the information provided by a DM, which will be used to trigger appropriate responses to these changes. In~\cite{shavarani2023detecting}, it was observed that while the proposed detection method successfully identifies relevant objectives, these objectives are sometimes replaced by irrelevant ones in subsequent iterations once they reach a local or global optimum and become fixed. In this study, we also investigate a method to prevent the algorithm from deactivating relevant objectives solely based on diminished correlation. This method can also be utilized for the early termination of the algorithm, thereby further reducing computational effort.

To validate the performance of the proposed detection model and examine the effects of preference drift and the designed responses, we performed a comprehensive experimental study. This study uses a diverse set of UFs to simulate a wide range of DM behaviors, tackle problems of varying dimensions and complexities, and evaluate multiple aspects of the algorithms. In addition, it evaluates various parameters for the interactive method and the associated detection and mitigation strategies. In addition, a sensitivity analysis is conducted to assess the impact of various parameters within the algorithm developed. This comprehensive analysis evaluates the robustness and reliability of the proposed detection method.
% \textcolor{red}{Another key contribution of this work is addressing the issues related to the significance of elicited preference information in the presence of preference drift. To provide reliable information based on the evolving nature of the DM's preferences, this study explores various strategies for resetting previously elicited preference information.}

Experimental results demonstrate that integrating dynamic detection methods and responsive mechanisms significantly enhances the adaptability and efficiency of the optimization process. Our findings show that these methods successfully refine the set of objectives, manage preference changes, and improve the relevance of solutions, thereby reducing computational effort and improving overall performance in dynamic optimization environments. This comprehensive analysis confirms the effectiveness of our approach in addressing real-world complexities and dynamic preference behavior in multi-objective optimization.
%Building on \cite{shavarani2023detecting} as the pilot study on detecting hidden and irrelevant objectives in iEMOAs, the present work examines the consequences of preference drift on the performance of the detection method. This study considers a range of preference drift scenarios, spanning from mild drifts to extreme changes in the decision maker's preferences. It is incontrovertible that the preference drift can raise doubts about the significance of elicited preference information in previous interactions. To solve related issues in this regard, this study also considers different options regarding resetting the previously elicited preference information in order to provide reliance on the accurate information based on the evolving preference nature of the DM. To validate the performance of the proposed detection model, this study employs various UFs to simulate diverse DMs exhibiting varied preference drift behaviors. Besides, detecting relevant objectives in such a dynamic preference behavior is examined using various feature selection methods, taking into account the varying dimensionalities, complexities, and structures of Pareto fronts across different problems. Ultimately, this study conducts a sensitivity analysis to assess the impact of various parameters within the developed algorithm.  

 %Experimental results show that the proposed method can almost always replace irrelevant objectives with relevant ones quickly and significantly improve the utility of the solutions found.

The remainder of the paper is structured as follows. Section~\ref{definitions} defines several key concepts that underlie this study. Section~\ref{background} provides a concise overview of previous efforts in objective reduction. In Section~\ref{methods}, the proposed method and its various iterations are expounded upon. Section~\ref{experiments} outlines the experimental setup. The results of the experiments are analyzed in Section~\ref{results}. Lastly, Section~\ref{conclusion} presents the conclusions drawn from the study and suggests avenues for future research.

\section{Definitions}\label{definitions}
This section provides a set of key definitions to ensure a clear understanding of the concepts discussed in this paper. For completeness reasons and due to the novelty of the problem setting, we will recap relevant definitions as defined in our previous work~\cite{shavarani2023detecting}. 

Without loss of generality, this study focuses on a minimization multi-objective optimization problem, which is commonly formulated as follows~\cite{battiti2010brain}:
\begin{equation}\label{eq:mo}
\text{Minimize} \quad f(\mathbf{x}) = { f_1(\mathbf{x}), \ldots, f_m(\mathbf{x}) }
\end{equation}
\begin{equation}
\text{subject to} \quad \mathbf{x} \in \Omega
\end{equation}

In this context, $\vec{x}\in \mathbb{R}^n$ is a solution vector comprising $n$ decision variables, and $\Omega \in \mathbb{R}^n$ represents the feasible region, defined by constraints on decision variables. For each $\vec{x}$, a set of $m$ numerical features, denoted as $F = {f_1, \dots, f_m}$, can be calculated and optimized as objective functions. The function \(f : \Omega \rightarrow \mathbb{R}^m\) maps each solution vector in \(\mathbb{R}^n\) to an objective vector in \(\mathbb{R}^m\).

\begin{definition}[Potential objectives]
The set of all features $F$, where $| F | = m$, is known as the \emph{potential} objectives.
\end{definition}

\begin{definition}[Active objectives]
Assume that due to modeling decisions or efficiency considerations, only a subset $\hat{F} \subseteq F$ ($\hat{m} = \abs{\hat{F}} \leq m$) of the potential objectives need to be minimized. The objectives in set $\hat{F}$ are called \emph{active}. Inactive objectives ($F \setminus \hat{F}$) are either evaluated but ignored or not evaluated at all. Inactive objectives do not participate in the optimization process.
\end{definition}

\begin{definition}[Objective evaluation]\label{def:objective_evaluation}
In the EMOA literature, computational cost is typically measured by \emph{solution evaluations}, where each evaluation includes the evaluation of all active objectives. However, since different solutions can be evaluated against varying subsets of objectives, we use \emph{objective evaluations} instead. Each evaluation of an objective $f_i$ for a solution $\vec{x}$ counts as one objective evaluation. Consequently, the cost of a solution evaluation is $\hat{m}$ objective evaluations.
\end{definition}

\begin{definition}[Domination~\cite{ZitKnoThi2008quality}]
A solution $\mathbf{x}$ is said to dominate another solution $\mathbf{y}$ ($\mathbf{x}, \mathbf{y} \in \Omega$) if $\mathbf{x}$ is at least as good as $\mathbf{y}$ in all objectives and strictly better in at least one objective.
\end{definition}

\begin{definition}[Pareto optimal~\cite{ZitKnoThi2008quality}]
A feasible solution $\vec{x}\in \Omega$ is Pareto optimal if there exists no other feasible solution $\vec{y}\in \Omega$ such that $\vec{y}$ dominates $\vec{x}$. When actively adjusting objectives during optimization by selecting subsets from potential options, Pareto optimality is determined solely by evaluating solutions against these dynamically chosen active objectives. The set of Pareto optimal solutions, which are mutually non-dominated, forms the \emph{Pareto set}. 
\end{definition}

\begin{definition}[Pareto front~\cite{ZitKnoThi2008quality}]
The mapping of the Pareto set on the objective space ($\hat{F}$) is called the Pareto front (PF).
\end{definition}

\begin{definition}[Redundant objectives~\cite{GalLeb1977ejor}]
An objective is deemed \emph{redundant} if removing it from the active objectives does not alter the set of Pareto optimal solutions.~\citet{SaxRayDeb2009constrained} expand this definition to include objectives that do not conflict with any non-redundant objective.
\end{definition}

\begin{definition}[Utility functions]
Based on utility theory~\cite{keeney1993decisions}, a utility function (UF) within the context of iEMOAs is defined as an unknown function that directs the DM's choices by effectively modeling her preferences. This UF is formally represented as $U\colon \mathbb{R}^{m} \to \mathbb{R}$, where the input is a vector-valued function $\vec{f}(\vec{x}) = (f_1(\vec{x}), \ldots, f_{m}(\vec{x}))$, representing the set $F$ of potential objectives~\cite{SteGar1991computational, AfsMieRui2021survey, fowgel2010ejor, PedTak2013emco, KokKar2010itdea}. Although $U$ accepts the values of all potential objective functions as input, it may selectively use only some of these values to determine its output. Non-ad-hoc interactive methods, which assume the existence of such a UF, stand in contrast to ad-hoc methods that operate without this assumption. This research adheres to a non-ad hoc approach, utilizing a UF to model preferences.
%In iEMOAs, the DM provides preference information, such as ranking a subset of solutions, to steer the algorithm toward her preferred solution(s). It is assumed that the DM can observe all potential objective values when comparing solutions. However, there may be a discrepancy between the objectives optimized by the iEMOA and those considered by the DM. 
%This discrepancy can be formally defined in non-ad-hoc interactive methods, which assume an unknown utility function (UF) guides the DM's decisions~\cite{SteGar1991computational,AfsMieRui2021survey, fowgel2010ejor, PedTak2013emco, KokKar2010itdea}. Ad-hoc methods assume no such UF exists~\cite{SteGar1991computational}. In alignment with the majority of studies in this domain, this research employs a non-ad-hoc approach, utilizing a UF to model preferences. Specifically, a UF is assumed to have the form $U\colon \mathbb{R}^{m} \to \mathbb{R}$, where the input is given by the vector-valued function $\vec{f}(\vec{x}) = (f_1(\vec{x}), \ldots, f_{m}(\vec{x}))$, representing the set $F$ of potential objectives. Although $U$ takes all potential objective values as input, it may not use all these values to compute its output.
%%%%%%%%%%%%%%%%%%%%%%%%%%%%%%%%%%%%%%%%%%%%%%%%%%%%
%
\end{definition}

%The above definitions are independent of the preferences of a human DM interacting with an EMOA. In the case of interactive EMOAs (iEMOAs), the DM provides preference information, e.g., by ranking a subset of solutions, to guide the algorithm towards the DM's most preferred solution. Let us assume the DM can observe the value of all potential objectives when comparing solutions.  For reasons explained in the introduction, there may exist a discrepancy between the active objectives being optimized by the iEMOA and the objectives considered by the DM when comparing solutions.

%We can formally define this discrepancy in the case of non-ad-hoc interactive methods, which assume there exists a utility function (UF) guiding the DM's decisions but unknown to the iEMOA~\cite{SteGar1991computational,AfsMieRui2021survey, fowgel2010ejor, PedTak2013emco, KokKar2010itdea}. Ad-hoc methods assume that no such UF exists~\cite{SteGar1991computational}.  Due to the popularity of UFs in modeling preferences, the vast majority of iEMOAs are non-ad-hoc methods, thus we focus on them in the remainder of the paper.  Without loss of generality,
% .
%we assume an UF of the form $U\colon \mathbb{R}^{m} \to \mathbb{R}$, whose input is the vector-valued function $\vec{f}(\vec{x})=(f_1(\vec{x}), \ldots, f_{m}(\vec{x}))$ with components being the set $F$ of potential objectives.
%
%
%Although $U$ receives as input the value of all potential objective functions, it may not use all those values to calculate its output.

\begin{definition}[Relevant \& irrelevant objectives]
An objective function $f_i \in F$ is deemed \emph{irrelevant} if its value does not influence the DM's UF. This means that if two solutions $\vec{x}$ and $\vec{y}$ in $\mathcal{X}$ yield identical values for all objectives except $f_i$, then they should also result in the same utility value. Formally, if $f_j(\vec{x}) = f_j(\vec{y})$ for all $f_j \in F \setminus {f_i}$, it follows that $U(\vec{f}(\vec{x})) = U(\vec{f}(\vec{y}))$. In this study, the set of objective functions relevant to the DM is denoted by $\Fdm \subseteq F$, while the set of irrelevant objectives is represented by $F \setminus \Fdm$.
\end{definition}

%\begin{definition}[Irrelevant objectives]
  %An objective  $f_i \in F$ is called \emph{irrelevant} if its value does not affect the value of the DM's UF. That is, any two solutions $\vec{x},\vec{y} \in \mathcal{X}$  with the same value in all potential objectives except $f_i$ should also have the same utility value, i.e., $f_j(\vec{x}) = f_j(\vec{y})$, $\forall f_j \in F\setminus\{f_i\}$  $\Rightarrow$  $U(\vec{f}(\vec{x})) = U(\vec{f}(\vec{y}))$. %\M.
%\end{definition}

%Hereafter, $\Fdm \subseteq F$ denotes the set of objective functions relevant to the DM, thus the set of irrelevant objectives is given by $F\setminus\Fdm$.

\begin{definition}[Hidden objectives]
An objective $f_i \in F$ is termed \emph{hidden} if it is relevant but not currently active, that is, $f_i \in \Fdm$ and $f_i \notin \hat{F}$. Hidden objectives can complicate iEMOA, as a solution that is non-dominated under the set of active objectives ($\hat{F}$) may be ranked lower than a dominated solution by the DM because of the discrepancy between the set of active objectives and those that are relevant to the DM ($\Fdm$). %This misalignment can cause suboptimal solutions to be selected, as the dominance criteria used by the algorithm might not reflect the full range of objectives that are ultimately relevant, leading to inaccurate assessments of solution quality.} %Hidden objectives can complicate iEMOA, as interactions with the DM can align with the dominance criterion for the objectives in $\Fdm$ but not for those in $\hat{F}$. 
If $\Fdm \subset \hat{F}$, there are no hidden objectives. However, iEMOA can optimize irrelevant objectives, which can be inefficient and costly, particularly if these evaluations are expensive. On the other hand, if $\Fdm = \hat{F}$, there are no hidden or irrelevant objectives and iEMOA effectively optimizes the objectives of interest to the DM.
\end{definition}

%\begin{definition}[Hidden objectives]
  %An objective $f_i \in F$ is \emph{hidden} if it is relevant but not (currently) active, i.e., $f_i \in \Fdm \land f_i \notin \hat{F}$. 
%
%\end{definition}

%Hidden objectives may confuse the iEMOA, since the interaction with the DM may be consistent with the dominance criterion for the objectives in $\Fdm$ but not for the objectives in $\hat{F}$.  If $\Fdm \subset \hat{F}$, then no hidden objectives exist, but the iEMOA is optimizing some irrelevant objectives, which makes the problem more challenging for the iEMOA and is wasteful if the evaluation of those objectives is expensive. Similarly,  if $\Fdm = \hat{F}$, then neither hidden nor irrelevant objectives exist, and the iEMOA is optimizing precisely the objectives that the DM cares about.

%\todo{IN LAZEME? MAN INO PARAPH NAKARDAM HANOOZ}In the rest of the paper, when considering benchmark problems and known UFs, we will assume for simplicity that irrelevant objectives are not a (trivial) function of relevant ones nor vice versa, so that the set of relevant objectives $\Fdm$, and, hence, irrelevant and hidden ones, can be inferred from the definition of the UF. In practice, the DM's UF is unknown and, in the case of black-box optimization, we may not know whether an objective is a function of other objectives, thus an objective is considered irrelevant if its value does not seem to influence the DM's decisions.

Based on the definitions provided, redundant objectives are defined by the problem's structure, while irrelevant and hidden objectives are defined from the DM's viewpoint. %\textcolor{red}{An irrelevant objective may or may not be redundant, but a redundant objective cannot be relevant unless the DM's preferences conflict with Pareto optimality. Conversely, a redundant objective might appear relevant if it correlates with a relevant one.}\todo{I'm not sure if this is true.} 
Although research exists on identifying and eliminating redundant objectives, there is a lack of studies on identifying irrelevant and hidden objectives in light of changes in the DM's preferences. This paper addresses this gap by proposing a method for detecting these objectives, detailed in Section~\ref{methods}.

%From the above definitions, it can be concluded that while \emph{redundant} objectives are determined based on the structure of the problem, \emph{irrelevant} and \emph{hidden} objectives are defined from the DM's perspective. An irrelevant objective may be redundant or not, however, a redundant objective cannot be relevant unless the DM's preferences are somehow inconsistent with Pareto optimality. On the other hand, a redundant objective may \emph{appear to be} relevant if it is correlated with a relevant objective. While there are studies on the detection and elimination of redundant objectives, which we review in the next section, there is no prior research on the identification of irrelevant and hidden objectives to the best of our knowledge. Our focus here is to fill this gap and we propose a method to tackle it in Section \ref{methods}.

\section{Background and Literature Review}\label{background}%Objective Reduction Approaches
A comprehensive literature review of methods for reducing the number of objectives has been conducted and is summarized in this section. Many of these studies use the term \emph{dimension} reduction to denote this concept. However, to avoid confusion with methods that reduce the number of decision variables~\cite{AllKno2010variables}, the term ``objective reduction'' is preferred.

Before discussing the research studies delving into objective reduction, it is essential to outline some key concepts related to multi-objective optimization. The primary goal in multi-objective problem-solving is to identify satisfactory non-dominated solutions from the PF. This process typically involves two distinct phases~\cite{miettinen2008introduction}: (1) the optimization phase, which aims to approximate the PF, and (2) the decision-making phase, focused on determining the DM's Most Preferred Solution (MPS). Depending on when the decision-making phase is applied, multi-objective optimization methods are classified into several classes~\cite{miettinen2008introduction}. \textit{A priori} methods require the DM to state their preferences in advance, often by assigning weights to different objectives. This can be challenging without knowing the possibilities and the complex nature of the data involved. On the other hand, \textit{a posteriori} methods involve the algorithm generating a representative subset of Pareto-optimal solutions, from which the DM can make a selection. These methods require the DM to choose from a potentially large pool of solutions, which can be labor-intensive or even impractical, especially in high-dimensional problems \cite{li2015many,purshouse2007evolutionary}.

Most of the research on objective reduction concentrates on selecting a subset of objectives \emph{a priori} to streamline the optimization process while preserving Pareto optimal solutions as much as possible. Early proposals~\citep{GalLeb1977ejor,Agrell1997ejor} imposed stringent assumptions about the problem structure that are impractical for real-world problems. More recent approaches~\cite{brozit2006dimensionality,SinIsaTap2011pareto} compromise on exactly identifying the correct subset of objectives and capturing the entire Pareto front to improve applicability.

%We have carried out a thorough literature review of methods for reducing the number of objectives, which we briefly summarize here.   
%Many of these studies use the term \emph{dimension} reduction to refer to the same concept. However, to avoid confusion with methods that reduce the number of decision variables~\cite{AllKno2010variables}, we use the term ``objective reduction''.

%Most of the studies on objective reduction focus on selecting \emph{a priori} a subset of objectives to facilitate the optimization process  while preserving Pareto optimal solutions as much as possible. Early proposals~\citep{GalLeb1977ejor,Agrell1997ejor} make strict assumptions about the problem structure that are impossible to meet for real-life problems. Recent approaches~\cite{brozit2006dimensionality,SinIsaTap2011pareto} sacrifice to exactly identify the correct subset of objectives and capture the entire PF in order to increase applicability.
%

Various methodologies aim to consolidate similar objectives into a single one. For example, harmonic levels~\cite{FreFleGui2013NonParametric} and aggregation trees~\cite{FreFleGui2015aggregation} are used to identify objectives that improve together without causing detriment to each other. Aggregation trees, in particular, are utilized \textit{a posteriori} to aid in the decision-making process. Principal component analysis (PCA) is also used to recognize correlated objectives that can be combined into one, either \textit{a posteriori} to facilitate decision making~\cite{LygMarFle2010Reduction} or during the optimization process to improve computational efficiency~\cite{SaxDurDebZha2013pca}. However,~\citet{CosOli2010biplotsReduction} demonstrated that objectives considered redundant by PCA might still be ``informative'', meaning that they contain trade-off information that could be lost if excluded.
%Other approaches identify similar objectives and recombine them into a single one. For example, harmonic levels \cite{FreFleGui2013NonParametric} and aggregation trees \cite{FreFleGui2015aggregation} are used to identify harmonious objectives (improvement of one objective does not lead to deterioration of the others). In particular, aggregation trees are used \textit{a posteriori} for facilitating the decision-making phase.
%
%
%Similarly, principal component analysis (PCA) has been used to identify correlated objectives that may be combined into a single objective\textit{ a posteriori} to facilitate decision-making~\cite{LygMarFle2010Reduction} or during the optimization process to increase computational efficiency~\cite{SaxDurDebZha2013pca}. However, \citet{CosOli2010biplotsReduction} have shown that objectives that are deemed redundant by PCA may be ``informative'', i.e.,  contain trade-off information that would be lost if omitted. 

Projection methods map all objectives into two or three dimensions for visualization purposes and are designed to assist in \textit{a posteriori} decision-making \cite{FieEve2013visualising}. However, these methods do not contribute to the optimization process itself and do not take the DM's preferences into account.
%Finally, projection methods map all objectives into two or three dimensions for visualization~\citep{FieEve2013visualising}. These methods aim to help decision-making \emph{a posteriori} (after optimization), however, they do not help the optimization process itself and do not consider the DM's preferences.

%ndle. 
% \end{comment}

%As mentioned earlier, having a comprehensive knowledge of the problem at hand required in \textit{a priori} methods, or the need for the labor-intensive analysis of a myriad number of solutions \textit{a posteriori} methods impose significant challenges. Interactive methods address these challenges %associated with \textit{a priori} and \textit{a posteriori} methods 
%by integrating the optimization and decision-making phases. 
As noted previously, \textit{ a priori} methods require an in-depth understanding of the problem beforehand, while \textit{a posteriori} methods require extensive analysis of numerous solutions, both of which present substantial challenges. Interactive methods overcome these obstacles by merging the optimization and decision-making phases. They achieve this by continuously engaging the DM during the optimization process and incorporating their preferences. This ongoing input directs the search to the most preferred areas of the PF~\cite{XinCheChe2018review,MOOINTEVO2008, shavarani2023interactive}. 
% \todo{in paragraph lazeme?}
% \textcolor{red}{Building on this foundation, it is crucial to distinguish between direct and indirect preference elicitation within interactive methods \cite{TomKad2020Indirect}. Direct preference elicitation necessitates that the DM specifies certain parameters of the preference model directly. These parameters can include the reference point (aspiration level/goal) (see, e.g., PBEA \cite{thimiekormol2009ec}, WASF-GA~\cite{RuiLuqMieSab2015}), reservation levels~\cite{GonSabRuiLuq2021Reservation}, and weights (see, e.g., R-NSGAII \cite{DebSun2006gecco}), among others. Conversely, indirect approaches require the DM to provide holistic judgments, which are generally less demanding and often take the form of exemplary decisions. In indirect methods, the DM does not need prior knowledge of the solution space and the optimization algorithm \cite{jacSis2001preference}. Indirect queries can include pairwise comparisons \cite{BatPas2010tec, BranCorrGreSlow2016ejor, TomKad2019iemoi, BranCorrGreSlow2016ejor, BenLerLus2020RIGA, TomKad2019emosor, BraGreSlo2010bpas}, selecting the best from a small subset of solutions \cite{KokKar2010itdea, fowgelkok2010interactive, TomKad2021ciemod}, accepting or rejecting a presented trade-off \cite{ZioWal1983interactive}, or ordering a subset of solutions \cite{sinmalkal2018convex}. Generally, iEMOAs that require the DM to rank a subset of solutions are referred to as ranking-based iEMOAs.}

The main objective of preference-based EMOAs is to identify solutions that align with the preferences of the DM by directing the search towards a specific area of interest. As the population converges towards the PF, a DM progressively gains deeper insights into the problem landscape through continual examination of newly generated solutions. Consequently, there exists a likelihood that DMs may revise their manifested preference information during the search process upon realizing the inadequacy of their initial preferences~\cite{lai2021empirical}. In theory, employing reference point methods could be viewed as a natural strategy
for accommodating preference drifts: the DM retains the freedom to adjust the reference point, thereby exploring novel regions of the Pareto front in response to alterations in preferences. However, the task of modifying the reference point becomes challenging for the DM when her preference model encompasses non-linear relationships between objectives. As the dimensionality of the problem increases, introducing a wide array of candidate directions for shifting the reference point, the cognitive demands become increasingly impractical~\cite{campigotto2010reactive}. 

A key advantage of interactive techniques is that they give the DM the opportunity to reconsider their choices, fix errors, and modify their preferences~\cite{thiele2009preference, sinha2011progressively}.~\citet{guo2019novel} have proposed a preference articulation methodology that entails uniform partitioning of preference regions based on the nadir point. Their method dynamically tracks evolving preferences to guide the evolutionary trajectory toward potentially viable regions. The main purpose of their study was twofold: to efficiently identify preferred solutions within a limited number of generations and to reduce the plethora of Pareto-optimal solutions presented to DM in \textit{a posteriori} evaluations.~\citet{gong2014interactive} introduced an iEMOA algorithm where the preferences of the DM
are expressed through the relative importance of objectives. Moreover, the DM has the capability to adjust these preferences interactively throughout the evolutionary process. Their study demonstrated that the frequency of preference adjustments influences the extent to which Pareto-optimal solutions align with evolving preferences, particularly when the number of generations is insufficient. 
As mentioned by~\citet{8789949}, a drastic change in preferences can slow down the search process or cause it to get `stuck' in a local optimum of a complex, multi-modal problem. To solve this issue, they improved responsiveness to preference changes using secondary population archives. When preferences change and the search is redirected to a new region of interest, an archive of previously found solutions is consulted, and solutions close to the new region are included in the current population.~\citet{battiti2010brain} utilized the support vector machine (SVM) classification algorithm to facilitate preference learning in a genetic technique called the Brain-Computer Evolutionary Multi-Objective Optimization algorithm (BC-EMOA). As mentioned by~\citet{campigotto2010reactive}, preference drift is effectively tackled by the BC-EMOA through a combination of strategies including implementing a discounting policy for outdated training examples during the learning phase and promoting diversification within the genetic population during the search phase.  

%Several studies have introduced machine learning (ML)-based techniques for iteratively learning user preferences. Although these techniques can be easily adapted to tackle preference drifts, these approaches are subject to various limitations (Batti 2010). The method proposed in [Sun1996] does not ensure the generation of Pareto optimal solutions, while the strategies developed in (Sun2000, Huang2005) produce a linear local approximation of user preferences and do not directly utilize the learned preference model to guide the search process. Moreover, in all these studies, feedback from the DM is communicated in terms of quantitative scores. 
%However, these limitations have been addressed by a genetic technique called the Brain-Computer Evolutionary Multi-Objective Optimization (BC-EMO) algorithm (Batti 2010). This method overcomes the challenges by employing pairwise preference supervision for learning the preference model, which significantly reduces the burden on the DM. Moreover, the algorithm utilizes the preference model directly to guide the search across the Pareto front. As mentioned by (Campigotto2010), preference drift is effectively tackled by the BC-EMO algorithm through a combination of strategies including implementing a discounting policy for outdated training examples during the learning phase and promoting diversification within the genetic population during the search phase.

\citet{shavarani2023detecting} were the first to integrate the DM's preference information within an iEMOA to identify the most relevant objectives to the DM. They utilized uni-variate feature selection and recursive feature elimination techniques to detect hidden and irrelevant objectives, drawing on preference information collected from the DM through a series of interactions. Their results demonstrated the method's efficiency in determining the set of objectives pertinent to the DM, effectively applying objective reduction while maintaining or even enhancing the utility of the DM's most preferred solution. However, the assumption of a fixed DM preference model limits the applicability of their method in realistic scenarios.  

The current study aims to extend~\cite{shavarani2023detecting} by incorporating the idea of preference drift through a range of different scenarios. To ensure accurate information that reflects the changing nature of the DM's preferences, this study investigates different strategies for resetting previously gathered preference information. 
Based on utility theory~\cite{keeney1993decisions}, which posits that a utility value influences the preferences of the DM when balancing trade-offs between conflicting objectives~\cite{steuer1991computational}, this study uses the Machine Decision Maker (MDM) introduced in~\cite{shavarani2022thesis} to validate the performance of the proposed detection model through a diverse set of UFs that simulate a wide range of DMs that exhibit different preference drift behaviors~\cite{shavarani2023benchmarking}.

\section{Methods}\label{methods}
This section outlines the methodology used in our study, starting with an overview of the detection of hidden and irrelevant objectives as proposed in~\cite{shavarani2023detecting}. Detailed formulations and discussions are available in the original study. We then elaborate on the modifications introduced to simulate and manage dynamic changes in the DM's preferences.

\subsection{Original Methodology}
The previous study used the BCEMOA Algorithm, as introduced in~\cite{BatPas2010tec}, and enhanced it with Hidden and Irrelevant Objective Detection (BCEMOA-HD). The algorithm is based on NSGA-II and involves several key steps:

\begin{enumerate}
\item \textbf{Initialization and Population Evolution:} The algorithm commenced with an initial population of randomly generated solutions, which are then evolved using the NSGA-II algorithm for a predefined number of generations ($gen_1$). 

\item \textbf{Interaction with the DM:} At each interaction step, a subset of solutions with size $N_exa$ was selected and presented to the DM. The DM ranked these solutions based on their perceived utility. This interaction allowed the DM to influence the optimization process by providing preferences that guided the search towards more desirable solutions.

\item \textbf{Dynamic Adjustment of Active Objectives:} After each interaction, univariate feature selection or recursive feature elimination methods were applied to assess the relevance of each objective. Objectives deemed irrelevant based on the DM's preferences were deactivated, while relevant objectives were activated. This focused the optimization on the most pertinent objectives, reducing computational complexity and improving alignment with the DM's preferences.

\item \textbf{Learning the UF:} The rankings provided by the DM, along with the values of detected relevant objectives, were used to train a Support Vector Machine (SVM) model. This model learned a UF ($U_{SVM}$) that predicted the DM's preferences, effectively capturing the DM's decision-making criteria.

\item \textbf{Guiding the Optimization Process:} The learned UF replaced the crowding distance metric in NSGA-II for subsequent generations. 
This adjustment directed the optimization process towards solutions that more closely matched the DM's preferences, ensuring that the search concentrated on the most promising regions of the solution space.
\end{enumerate}

\subsection{Modified Methodology}
We extend our previous methodology~\cite{shavarani2023detecting} in several aspects to account for dynamic changes in the preferences of the DM including (i)~simulating preference changes, (ii)~detecting these changes, and (iii)~dynamically adjusting the optimization process.

\subsubsection{Simulation of Preference Changes}\label{simulation_of_preference_changes}
As in the original study, the behavior of the DM is simulated using different UFs. To simulate different degrees of preference change we modify the UFs by changing the weights of objective functions. To do so, we follow and extend the method proposed by~\citet{Ste1996jmcda} to impose the preference dependency between objectives. 
Let $C = \{c_1, c_2, \dotsc, c_{m-q}\}$ be the index set of relevant objectives in random order. Each pair of objectives $(z_{c_k}, z_{c_{k+1}})$, where $k = 1, 3, 5, \dots$, $k < |C|$, contain an irrelevant and relevant objective and is modified as follows:

\begin{equation}\label{eq:mixing}
\begin{aligned}
  \tilde{z}_{c_k} &= (1-\gamma)z_{c_k} + \gamma z_{c_{k+1}}, \text{and} \\
  \tilde{z}_{c_{k+1}} &= (1-\gamma)z_{c_{k+1}} + \gamma z_{c_k}
\end{aligned}
\end{equation}

Here, $\gamma \in [0, 1]$ is a parameter that controls the intensity of the dependency between the objectives. A higher $\gamma$ leads to a more pronounced change in preference. When $\gamma = 0$, no preference change occurs, while $\gamma = 1$ indicates a complete reversal in the roles of relevant and irrelevant objectives. The adjustment of objective weights occurs only during interactions with the DM and does not impact the optimization process itself.

%where $\gamma \in [0, 1]$ is a parameter that controls the intensity of the dependency between the objectives. The higher the $\gamma$, the more severe the preference change. When $\gamma$ is set to zero, there is no preference change, while setting it to one indicates switching the role of relevant and irrelevant objectives. The changing of objective weights occurs exclusively during interactions with the DM and does not affect the optimization process itself.  

    %\item Moderate: In this scenario, we deactivate an irrelevant objective and activate a relevant one by assigning the place of the relevant objective in the utility value to the irrelevant one.
    %\item Severe: In this extreme scenario, all relevant objectives are treated as irrelevant, and all irrelevant objectives become relevant. This transformation is accomplished by setting the weights of all relevant objectives to zero while assigning equal weights to all irrelevant objectives participating in the utility calculation.

\subsubsection{Detection of Preference Changes}
In each interaction, a mechanism is employed to detect potential changes in the DM's preferences. This is achieved by comparing the detected relevant objectives, based on their ranking provided in the current interaction, to the current set of active objectives. If a change is detected, it suggests that a moderate or severe preference change has occurred. Detection of moderate and severe preference changes is crucial because it requires updating the set of active objectives and potentially adjusting the population.

In such cases, methods such as archive preservation or data injection can be used to help the algorithm escape from the local optimum to which it may have been directed based on the previous set of objectives. These mechanisms are not discussed in this study due to page limitations.

\subsubsection{Elimination of Outdated Preference Data}

When a change in preferences is detected, the algorithm initiates a process to discard outdated and conflicting preference data. This ensures that the optimization process remains aligned with the DM's current priorities. To investigate the effects of eliminating conflicting preferences elicited in previous interactions, we implement a parameter called "reset." This parameter can be set to either $None$, $fixed$, or $dynamic$ to dictate the behavior of the algorithm regarding whether to eliminate or retain the conflicting data.

Setting the ``reset'' parameter to $dynamic$ directs the algorithm to discard conflicting preferences when a change in preferences is detected. Conversely, setting it to $None$ retains the conflicting preferences, allowing them to potentially be considered alongside new preferences in future iterations.  If the ``reset'' parameter is set to $fixed$, the algorithm exclusively uses the most recent preference data obtained from the latest interaction with the DM. This procedure is outlined in Algorithm \ref{reset}.

\begin{algorithm}
\caption{refine\_pref($pref$, $new\_pref$, $update\_needed$, $reset$)}
\label{reset}
\begin{algorithmic}[1]

\IF{$reset = $ \textbf{None}}
    \STATE \textbf{Return} $pref$ + $new\_pref$
\ELSIF{$reset$ = \textbf{fixed}}
    \STATE \textbf{Return} $new\_pref$
\ELSIF{$reset$ = \textbf{dynamic}}
    \IF{$update\_needed$}
        \STATE \textbf{Return} $new\_pref$
    \ELSE
        \STATE \textbf{Return} $pref + new\_pref$
    \ENDIF
\ENDIF
\end{algorithmic}
\end{algorithm}

\subsubsection{Interactive Process with Preference Detection}

The modified interactive process is as follows:
\begin{enumerate}[label=\Roman*.]
    \item Initialize the population and evolve solutions using NSGA-II.
    \item At each interaction step, present a subset of solutions to the DM for ranking.
    \item Detect hidden/irrelevant objectives and update the population if necessary.
    \item Detect any changes in the DM's preferences by contrasting the relevant objectives based on the latest preference information and the current one. If a change is detected, eliminate outdated preference data and update active objectives.
    \item Use the updated preference data to estimate a utility value.
    \item Guide the optimization process using the updated UF and active objectives.
\end{enumerate}

In~\cite{shavarani2023detecting} it was noted that relevant objectives can become trapped in local or global optima, causing their values to remain fixed. As the correlation of a fixed vector with the ranks decreases to zero, these objectives may mistakenly be detected as irrelevant. To mitigate this issue, one approach is to introduce noise to the objective values that remain fixed across all solutions presented in an interaction. This adjustment aims to prevent such scenarios from occurring, ensuring that relevant objectives are correctly identified.
In this study, we will investigate the effectiveness of introducing noise to prevent misclassification of fixed objective values as irrelevant. 
%Furthermore, in the original study~\cite{shavarani2023detecting}, it was not possible to terminate the optimization when no further improvements were achieved in the estimated UF. Instead, a new set of objectives would typically be detected. In contrast, it is now possible to implement a termination criterion based on the UF, which terminates the optimization process when no additional improvements are observed.
These advancements allow for more efficient optimization by avoiding unnecessary computations when further improvements in the UF are unlikely to occur. The iEMOA framework equipped with mechanisms for detecting relevant objectives and dynamically adjusting to the evolving preferences of the decision maker is detailed in Algorithms \ref{BCEMOA HD}.

\begin{algorithm}
\caption{iEMOA framework for dynamic preference adaptation}
\label{BCEMOA HD}
\begin{algorithmic}[1]
\STATE \textbf{Input} ($gen\_i$, $gen\_1$, $interactions$, $sample\_size$, $reset$, $add\_noise$, $pref\_learning$)

\STATE $Pop$ $\leftarrow$ Initialize population
\STATE $interaction$ $\leftarrow$ $0$
\STATE $pref$ $\gets \{\}$ 
\STATE $Pop \leftarrow$ evolve $Pop$ with NSGA2 for $gen\_1$ generations
\WHILE{$interaction \leq interactions$}
    \STATE $interaction \gets interaction + 1$
    \IF{$pref\_learning$ is enabled}
        \STATE $new\_pref \gets$ $sample\_size$ randomly selected solutions ranked by the DM %get\_preferences(Pop, pref, sample\_size)$
        \IF{$add\_noise$}
            \STATE Add noise to the fixed values in $new\_pref$
        \ENDIF
        \STATE $\hat{F}_{new} \gets \text{identify relevant objectives based on } new\_pref$ \{As described in \cite{shavarani2023detecting}\}
        \STATE $update\_needed \gets False$ 
        \IF{$\hat{F}_{new} \not\subseteq \hat{F}$}
            \STATE Update the set of active objectives
            \STATE $update\_needed \gets True$ 
        \ENDIF
        \STATE $pref \gets refine\_pref(pref, new\_pref, update\_needed, reset)$
        \STATE $ranking\_criteria \gets learn(pref)$ \{train the estimator on the $pref$\}
    \ELSE
        \STATE $ranking\_criteria \gets \textbf{real UF} $ \{The ideal condition with no learning\}
    \ENDIF
    \STATE $NSGA2 \gets$ replace the crowding distance with $ranking\_criteria$
    \STATE $Pop \leftarrow$ evolve $Pop$ with NSGA2 for $gen\_i$ generations
\ENDWHILE
\STATE return $Pop$
\end{algorithmic}
\end{algorithm}

\section{Experimental Setup} 
We conduct a comprehensive set of experiments to evaluate the performance of the algorithm concerning several key aspects, including the detection of hidden and irrelevant objectives, handling the dynamic nature of DM's preferences, examining the effects of different levels of preference changes, managing potential changes in relevant objectives through the proposed noise disturbance, and assessing the impact of objective reduction. To achieve this, we include experiments on various benchmark problems with different dimensions and levels of difficulty.

We utilize two renowned numerical benchmark problem sets to evaluate our method: multi-objective NK landscape problems with correlated objectives ($\rho$MNK)~\cite{VerLieJou2013ejor} and DTLZ problems~\cite{DebThiLau2005dtlz}, considering cases with $m \in {4, 10, 20}$ objectives.

The $\rho$MNK problems enable us to analyze how the correlations among the objectives and the smoothness of the landscape affect the performance of the proposed method. We examine $\rho$MNK instances with varying correlation values among objectives ($\rho \in {0, 0.9}$), adhering to the constraint $\rho \geq -1/(m-1)$~\cite{VerLieJou2013ejor}. The dimension of the solution space is determined by $n = (round(m/10)+1)*10$~\cite{shavarani2023detecting}. We also explore different values for the parameter $K$, which dictates landscape smoothness: $K \in {1, 8}$ for problems with 4 objectives and $K \in {1, 15}$ for many-objective problems (10 and 20 objectives), ensuring $K < n$~\cite{VerLieJou2013ejor}. Higher values of $K$ correspond to more rugged fitness landscapes. The variable $n$ is fixed at $10$ for $m=4$, 20 for $m=10$, and 30 for $m=20$ in $\rho$MNK problems.

Adhering to the original BCEMOA experiments~\cite{BatPas2010tec} and studies on hidden objectives and objective reduction~\cite{shavarani2023detecting,BroZit2007hypervolumeReduction}, we focus on DTLZ1, DTLZ2, and DTLZ7 from the DTLZ test suite. DTLZ1, with its $11^k-1$ local Pareto-optimal fronts, tests the algorithm's capacity to manage multiple local attractors. DTLZ2, featuring a concave Pareto front, and DTLZ7, with $2^{m-1}$ disconnected Pareto-optimal regions, are employed to evaluate solution diversity and algorithm performance in disconnected feasible spaces.

As recommended in~\cite{DebThiLau2005dtlz}, the dimension of the decision space ($n$) is set to $m+4$ for DTLZ1, $m+9$ for DTLZ2, and $m+19$ for DTLZ7. In DTLZ problems, optimizing a subset of objectives leads to the optimization of the remaining objectives. To increase the complexity of the problem and prevent the Pareto front (PF) from collapsing to a single point when projected to $k < m$ objectives, we follow the approach in~\cite{BroZit2007hypervolumeReduction}. Specifically, we map $x_i$ to $x_i/2 + 0.25$ for DTLZ2 and constrain $x_i$ within $[0.25, 0.75]$ for DTLZ1, as suggested in~\cite{BatPas2010tec}. These modifications are not necessary for DTLZ7, as it does not collapse to a single point.

\subsection{Simulating Active, Inactive, Hidden, and Irrelevant Objectives}\label{simul} % with hidden objectives}
Simulating active and inactive objectives is accomplished by expanding existing benchmark problems as follows: Given a problem with $m = \abs{F}$ potential objectives, we augment it with $\vec{d} \subseteq [1, m]$, an ordered index set of active objectives such that \mbox{$i < j \to d_i < d_j$}, indicating which objectives are active (optimized), i.e., \mbox{$i \in \vec{d}$} iff \mbox{$f_i \in \hat{F} \subseteq F$}, where $f_i$ is the $i^\text{th}$ potential objective function. Specifically, for a solution $\vec{x}$ with an objective vector $\vec{f}(\vec{x}) = (f_1(\vec{x}), \dotsc, f_m(\vec{x}))$, the optimizer only considers $\vec{\hat{f}}(\vec{x}) = (f_{d_1}(\vec{x}), \dotsc, f_{d_{\hat{m}}}(\vec{x}))$ and can modify the set of active objectives by changing the indices in $\vec{d}$.

In contrast, feature selection methods and the DM have access to $\vec{f}(\vec{x})$. Specifically, when evaluating a solution $\vec{x}$, the simulated DM evaluates $U(\vec{f}(\vec{x}))$, where $U$ is the UF reflecting the preferences of the DM.

To simulate an irrelevant objective, we set the objective's weight in the utility function ($UF$) to zero, while assigning positive weights to the relevant objectives. For simplicity, we assume that initially there are only two relevant objectives at the start of the optimization process. Therefore, the UF is constructed with all objective weights set to zero, except for the two deemed relevant. It is important to note that the optimizer, detection, and reduction methods do not have access to this UF; they are only used to simulate the DM's responses to queries. When selecting irrelevant objectives for the experiments, it is important to note that projecting the PF onto lower dimensions can cause it to collapse into a single point for some problems. This phenomenon is observed in DTLZ problems, even when the problem is bounded~\cite{BroZit2007hypervolumeReduction}. Therefore, careful attention must be paid when using these problems to simulate active and inactive objectives. For example, in DTLZ7, the PF collapses to a single point when the first two objectives are active and the remaining ones are inactive. Upon a detailed examination, the first and fourth objectives are designated as relevant for DTLZ problems ($\vec{c}={1,4}$). For $\rho$MNK problems, the first two objectives are chosen as relevant ($\vec{c}={1,2}$). For a $\rho$MNK problem with four objectives, where $\Fdm = {f_1, f_2}$ and an initial $\vec{d} = {2, 4 }$, we observe that $f_1$ is a hidden objective (relevant but not optimized), $f_2$ is both relevant and optimized, $f_3$ is irrelevant and not optimized, and $f_4$ is irrelevant and optimized.

To simulate different decision-making behaviors, different UFs are used.
%%%%%%%%%%%%%%%%%%%%%%%%%%%%%%%%
The constructed $UF$ is used to simulate DM's feedback within the MDM framework proposed in~\cite{LopKno2015emo}.
We consider the following Tchebychef UF:
 \begin{equation}\label{eq:tch}
    U_\text{tch}(\vec{f}) = \max_{i \in \vec{c}}w_i|f_i-f_i^*|
\end{equation}

with $\vec{0}$ as the ideal point $\vec{f}^*$.
The weights $w_i$ of irrelevant objectives ($i \notin \vec{c}$) are set to zero while the weights of relevant objectives were manually selected for each problem so that the most preferred solution is away from the corner points as far as possible.
Quadratic UF is also used as formulated in Eq~\ref{eq:quad}. The specific weights used in each UF for the respective problems are detailed at \texttt{www.to-be-updated-after-peer-review.com}  \textit{(please note that due to the double-blind review process, the link will be published after the paper is accepted)}.
 \begin{equation}\label{eq:quad}
    U_\text{Quad}(\vec{f}) = \sum_{i \in \vec{c}}{(w_i|f_i-f_i^*|)^2}
\end{equation}

It should be noted that we chose to limit the levels of certain parameters, as well as the number of problems and utility functions, given that the current parameter combinations already exceed $100,000$. With $40$ repetitions required for each configuration, this results in over $4$ million executions. Exhaustively exploring all possible combinations would have imposed an impractical computational burden on our study.

% Although the UFs in Eqs.~\eqref{eq:1}--\eqref{eq:tch} are designed for minimization, we reverse and
% scale all utility values in the experiments to the range $[0, 1]$, such that $1$ corresponds to the best utility
% value and $0$ to the worst one, for consistency with multi-attribute utility theory~\cite{keeRai1993decisions}.

\subsection{Evaluation of Results}
To comprehensively evaluate the performance of the proposed algorithm, we performed experiments using various combinations of key parameters. This approach allows us to analyze how different settings impact the effectiveness and robustness of the algorithm across diverse scenarios. In addition to parameters such as the number of objectives, the degree of correlation among objectives, and the smoothness of the fitness landscape, we examine different settings related to the algorithm itself and different capabilities such as detection, reduction, and the extent of preference changes:
\begin{enumerate}
\item \textit{Learning $\in [\text{True, False}]$}: When learning is set to \textit{False}, the algorithm operates as a non-interactive EMOA without any interaction. In contrast, when learning is \textit{True}, the system engages with the DM to learn their preferences. Regardless of the learning setting, there is no detection of relevant or irrelevant objectives, and the set of active objectives remains unchanged.
\item \textit{Detection $\in [\text{True, False}]$}: When detection is set to \textit{True}, the algorithm will identify and update relevant and irrelevant objectives after each interaction, dynamically adjusting the set of active objectives. If learning is set to \textit{False}, detection is automatically disabled and set to \textit{False} as well.

\item \textit{Reduction $\in [\text{True, False}]$}: As discussed earlier, detection can be performed with the number of objectives either fixed or variable. The latter case is achieved by setting Reduction to \textit{True}, where initially all objectives are active, but the algorithm can subsequently reduce the number of objectives to as few as two, based on their relevance. We evaluate various threshold values ($\tau$) to determine their impact on activating the relevant objective when the reduction is set to True.

\item \textit{Noise $\in [\text{True, False}]$}: In~\cite{shavarani2023detecting}, it was noted that relevant objectives could be mistakenly identified as irrelevant if they become fixed in local or global optima, causing their correlation with ranks to drop. Introducing noise to fixed objective values across all solutions presented to the DM can prevent this misidentification. The option to add noise to fixed objective values can be enabled or disabled by setting the noise option to \textit{True} or \textit{False}, respectively. The $Noise$ is automatically set to \textit{False} when detection is \textit{False}.
\item \textit{ $\gamma \in [\text{0, 0.25, 0.5, 0.75, 1}]$}: This parameter specifies the extent of change in the DM's preferences.
\item \textit{Preference Change}: In this set of experiments, a single change in the DM's preferences is introduced during the third interaction. The intensity of this preference change is regulated by the parameter $\gamma$.
\item \textit{Detection Method $\in [\text{None, Recursive, Univariate}]$}: Specifies the method used for detecting relevant and irrelevant objectives.

\item \textit{Preference Reset $\in [0, \text{Fixed, Dynamic}]$}: This parameter dictates the strategy for resetting preferences during the optimization process. A value of 0 indicates no reset. When set to 'Fixed', preferences are reset at each interaction. When set to 'Dynamic', preferences are reset only when a change in preferences is detected, specifically if the set of identified objectives is not a subset of the current active objectives.

\end{enumerate}

\subsubsection*{Parameter Settings}
All variants of BCEMOA adhere to the parameter settings outlined in the original paper~\cite{BatPas2010tec}, including those for the SVM learning model. Specifically, the total number of generations is set to $500$, and \(\Nexa = 5\) solutions are presented to the DM at each interaction. In BCEMOA, the NSGA-II algorithm employs a population size of $100$ and generates $100$ new solutions per generation. It operates for \(gen_1 = 200\) generations before the first interaction and continues with \(gen_i = 30\) generations between subsequent interactions. The total number of generations after the final interaction is computed as \(500 - gen_1 - gen_i(\Ninteractions - 1)\). Therefore, varying the number of interactions (\(\Ninteractions\)) does not affect the total number of generations.

We conducted experiments with $9$ interactions. For \(\rho\)MNK problems, we examined the impact of different levels of correlation (\(\rho\)) and ruggedness (\(K\)). Each configuration was executed $40$ times with different random seeds to ensure robustness in the results. 

\subsubsection*{Implementations}
The algorithms, DM, and $\rho MNK$ problems are implemented in Python 3.7.6. The NSGA-II algorithm used in BCEMOA and the DTLZ benchmarks is sourced from the Pygmo library version 2.16.0~\cite{BisIzzYam2010:pagmo}. For uni-variate feature selection and Recursive Feature Elimination (RFE), we utilize Scikit-learn version 0.23.1 (\url{http://scikit-learn.org/}). To encourage further research, we have made our code publicly available at \texttt{www.to-be-updated-after-peer-review.com}.

\section{Experimental Results} \label{results}
The results section provides a detailed analysis of the experiments conducted to evaluate the proposed dynamic approach for detecting hidden and irrelevant objectives in iEMOAs. The analysis focuses on several key aspects, including the effect of different detection methods, the impact of noise addition, and the behavior of the algorithm under varying preference change intensities. The results are detailed through a series of figures and tables, discussed in this section. The reported $p$-values are obtained from ANOVA tests conducted at a significance level of $0.05$ to assess statistical significance.

\subsection{Impact of Detection Methods and Reduction} 
The effectiveness of various detection methods (univariate and recursive feature elimination) and their interaction with the reduction method are evaluated across multiple interactions. The results are presented in Table~\ref{tab:learning_detection_reduction}, which shows the performance of the algorithm for different settings, including when learning is turned off. Deactivating the learning process (first row of Table~\ref{tab:learning_detection_reduction}) represents an ideal scenario in which the algorithm directly optimizes the UF. Although this situation is not realistic, it serves as a baseline for assessing the algorithm's performance.

Table~\ref{tab:learning_detection_reduction} shows that when the detection method is active, the algorithm generally performs better. With the reduction method enabled, all objectives are active from the start, leading to suboptimal utility values before the first interaction. However, as the algorithm progresses, it successfully improves the desirability of the solutions over successive interactions. Ultimately, the final results are more favorable compared to cases where the reduction method is disabled. This behavior mirrors the ideal scenario where learning is deactivated, even surpassing it in the last interaction. This improvement can be attributed to the dynamic nature of the reduction variant and its higher exploration rate.

\begin{table*}
\caption{The effect of different detection methods on the performance of the algorithm across 9 interactions, with interaction O indicating the utility of the best solution found before the first interaction. 
}
    \centering
    \begin{tabular}{lllrrrrrrrrrr}
\toprule
 &  & interaction & 0 & 1 & 2 & 3 & 4 & 5 & 6 & 7 & 8 & 9 \\
 \cline{3-13}
learning & detection & reduction &  &  &  &  &  &  &  &  &  &  \\
\midrule
0 & None & 0 & 0.588 & 0.891 & 0.891 & 0.891 & 0.891 & 0.891 & 0.891 & 0.891 & 0.891 & 0.891 \\
\cline{1-13} \cline{2-13}
\multirow[t]{5}{*}{1} & None & 0 & 0.786 & 0.833 & 0.836 & 0.841 & 0.845 & 0.848 & 0.847 & 0.848 & 0.847 & 0.871 \\
\cline{2-13}
 & \multirow[t]{2}{*}{recursive} & 0 & 0.786 & 0.834 & 0.835 & 0.838 & 0.853 & 0.854 & 0.855 & 0.856 & 0.856 & 0.880 \\
 &  & 1 & 0.581 & 0.734 & 0.740 & 0.743 & 0.825 & 0.829 & 0.834 & 0.834 & 0.836 & 0.886 \\
\cline{2-13}
 & \multirow[t]{2}{*}{univariate} & 0 & 0.785 & 0.830 & 0.830 & 0.832 & 0.853 & 0.853 & 0.854 & 0.855 & 0.856 & 0.883 \\
 &  & 1 & 0.583 & 0.760 & 0.744 & 0.759 & 0.833 & 0.835 & 0.840 & 0.842 & 0.844 & 0.893 \\
\bottomrule
\end{tabular}

\label{tab:learning_detection_reduction}
\end{table*}

Similar results are observed in Table~\ref{tab:problem_detection_reduction}, which presents a comparative analysis of the performance of the reduction method in different types of problems, including DTLZ1, DTLZ2, DTLZ7, and $\rho MNK$. This table highlights how the reduction method enhances the utility values across different interactions and problem types. In particular, the table shows that the reduction method ultimately leads to an increase in the desirability of solutions, regardless of the type of problem. This underscores the general applicability and effectiveness of the reduction method in improving optimization outcomes.
\begin{table*}
\caption{The performance of the reduction method across different problems. The desirability of solutions increases with the application of the reduction.}
    \centering
    \begin{tabular}{llrrrrrrrrrr}
\toprule
 & interaction & 0 & 1 & 2 & 3 & 4 & 5 & 6 & 7 & 8 & 9 \\
 \cline{2-12}
prob & reduction &  &  &  &  &  &  &  &  &  &  \\
\midrule
\multirow[t]{2}{*}{1} & 0 & 0.979 & 0.985 & 0.983 & 0.986 & 0.982 & 0.983 & 0.983 & 0.984 & 0.983 & 0.985 \\
 & 1 & 0.857 & 0.972 & 0.964 & 0.956 & 0.980 & 0.976 & 0.976 & 0.978 & 0.978 & 0.990 \\
\cline{1-12}
\multirow[t]{2}{*}{2} & 0 & 0.970 & 0.986 & 0.981 & 0.981 & 0.984 & 0.985 & 0.985 & 0.985 & 0.985 & 0.988 \\
 & 1 & 0.658 & 0.954 & 0.927 & 0.945 & 0.968 & 0.966 & 0.972 & 0.969 & 0.970 & 0.989 \\
\cline{1-12}
\multirow[t]{2}{*}{7} & 0 & 0.965 & 0.977 & 0.975 & 0.977 & 0.878 & 0.876 & 0.879 & 0.881 & 0.884 & 0.907 \\
 & 1 & 0.583 & 0.826 & 0.822 & 0.821 & 0.863 & 0.864 & 0.875 & 0.874 & 0.876 & 0.948 \\
\cline{1-12}
\multirow[t]{2}{*}{$\rho MNK$} & 0 & 0.642 & 0.721 & 0.724 & 0.729 & 0.780 & 0.781 & 0.781 & 0.783 & 0.783 & 0.819 \\
 & 1 & 0.494 & 0.619 & 0.620 & 0.634 & 0.748 & 0.754 & 0.758 & 0.761 & 0.764 & 0.824 \\
\bottomrule
\end{tabular}
\label{tab:problem_detection_reduction}
\end{table*}

Another interesting observation is the effect of the utility function on the performance of the reduction method. While for Tchebychef UF the reduction improves the utility value of the final solution by $20\%$, this improvement goes beyond $70\%$ when quadratic utility value is used. The same results are observed when investigating the effect of the UF across different detection methods.

\subsection{Dealing with Preference Change}
This section investigates the impact of preference changes simulated in the third interaction and evaluates the effectiveness of the proposed mechanisms in addressing these changes. As part of a pilot study exploring these scenarios and as discussed in Section~\ref{methods}, three variants are considered to refine the preference information elicited. In the ``fixed'' method, the algorithm relies solely on the latest preference information, discarding all previous data. The ``dynamic'' method takes a heuristic approach, resetting the preference information only when the set of detected relevant objectives is not a subset of currently active objectives. Lastly, the ``None'' method represents a scenario where the algorithm takes no specific action to handle potential preference changes. It is important to note that the algorithm is not informed about when or to what extent the DM's preferences change, if any. This approach is intended to simulate real-life conditions as closely as possible.

\begin{figure}[h]
    \centering
    \includegraphics[width=\columnwidth]{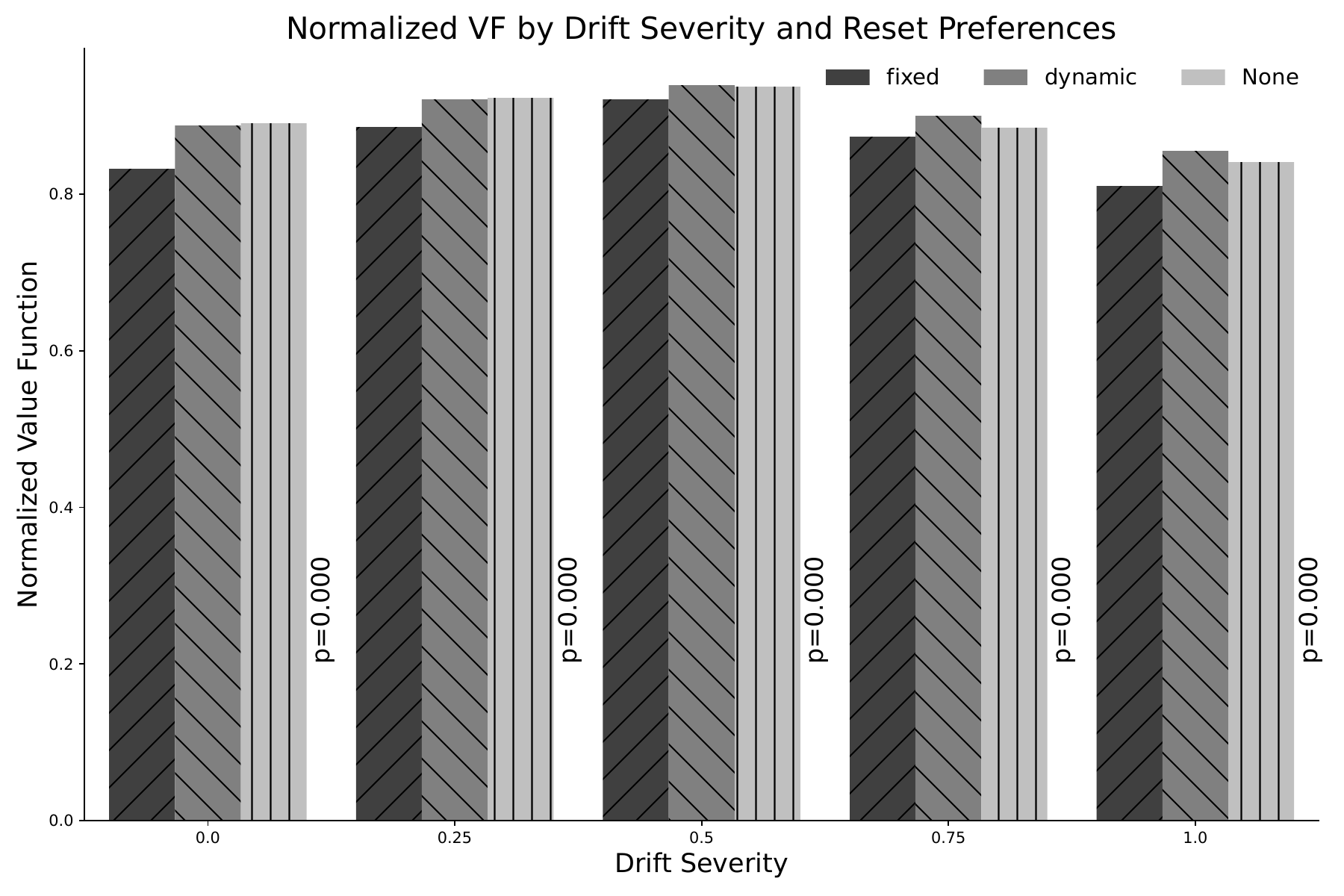}
    \caption{The utility values of the final solutions returned by the algorithm across different levels of drift severity. }
    \label{fig:vf_by_intensity_reset}
\end{figure}

Figure~\ref{fig:vf_by_intensity_reset} visualizes the utility values of the final solutions returned by the algorithm at different levels of preference change and reset options. The graph highlights the superior adaptability of the dynamic method compared to other approaches as the intensity of the preference changes increases. The dynamic reset method's capacity to sustain high performance, even under severe conditions, demonstrates its robustness and reliability. In contrast, the fixed reset option, which eliminates preference information after each interaction, consistently produces inferior results in all cases.

\begin{figure*}[h]
    \centering
    \includegraphics[width=\textwidth]{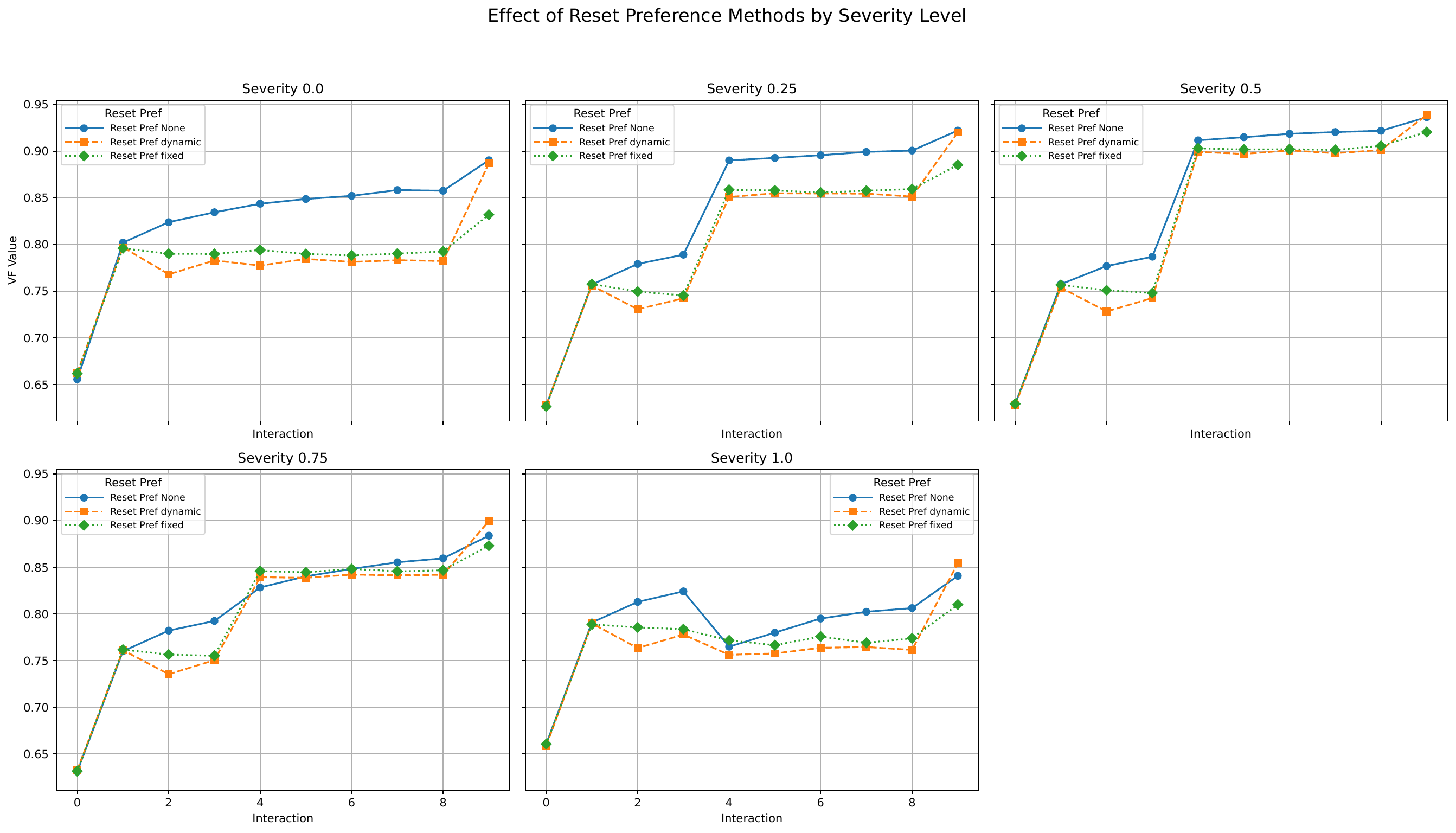}
    \caption{The effects of different reset preference methods on algorithm performance under varying intensities of preference change.  }
    \label{fig:severity_reset}
\end{figure*}

Figure~\ref{fig:severity_reset} offers an in-depth comparison of how various reset preference methods manage preference changes across a range of interactions and levels of reference drift intensity. At a severity level of 0.0, where no preference change is present, there is a significant performance disparity between the ``dynamic'' and ``None'' reset methods. However, as the severity of the preference changes increases, this performance gap narrows. In particular, when the severity reaches or exceeds 0.5, the difference diminishes, and the ``dynamic'' method even outperforms the ``None'' method during early interactions at a severity level of 0.75. Across all scenarios, the final solutions generated by the algorithms demonstrate improved desirability when the ``dynamic'' method is employed, highlighting its superior effectiveness in handling more substantial preference changes.
%The role of different reset preference methods in managing preference changes is examined over different interactions and under various reference drift intensities in Figure~\ref{fig:severity_reset}. While there is a large gap between "dynamic" and "None" when there is no preference change (Severity 0.0), their performance gets closer when increasing the preference change severity, even having superior results for the "dynamic" reset preference option in cases with severity more than or equal 0.5.   

%As the intensity of preference change increases, the dynamic reset method shows a clear improvement in performance at the last interaction, further validating its adaptability and robustness under more severe conditions.

\subsection{Effect of Adding Noise} 
Adding noise to fixed objective values would prevent misclassification of relevant objectives that become fixed in local or global optima. The amount of noise that is added to the objective functions in case of trapping in local optima is set to $1\%$  of the value of that objective function in that interaction.   %\textcolor{red}{Thus, this section explores the impact of noise addition across different problems.}%.

\begin{figure}[h]
    \centering
    \includegraphics[width=\columnwidth]{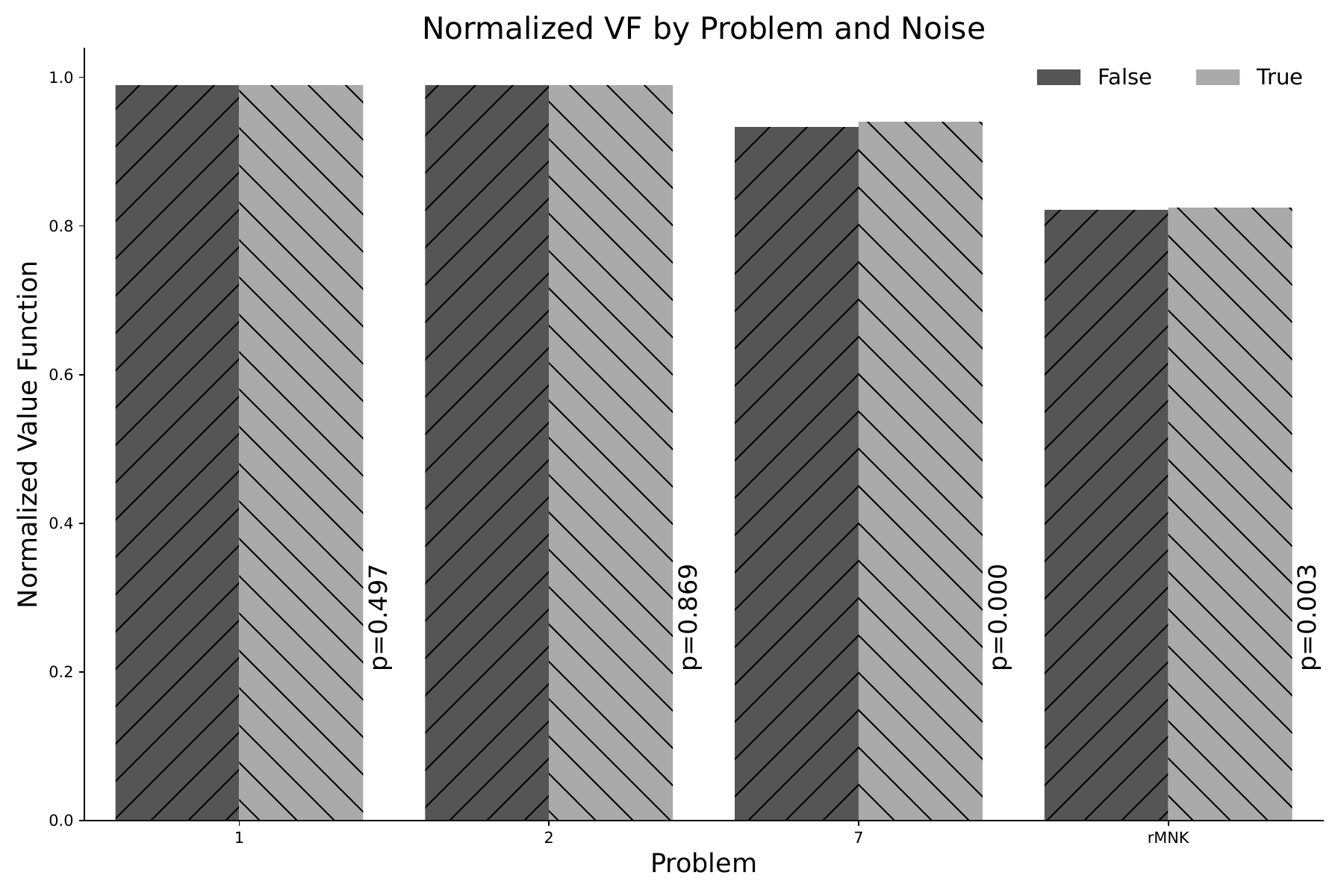}
    \caption{The impact of noise on the quality of results across various problems %This figure illustrates how adding noise to objective values, particularly when these values are fixed, enhances the robustness and desirability of the final results produced by the algorithm. The improvement is especially pronounced in the cases of the DTLZ 2 and $\rho MNK$ problems, highlighting the benefits of introducing variability in these scenarios. The analysis focuses on the outcomes observed after the $9^{th}$ interaction, demonstrating the long-term impact of noise addition on performance stability. 
    }
    \label{fig:prob_noise}
\end{figure}

The impact of noise addition is explored across different problems, as shown in Figure~\ref{fig:prob_noise}. This figure demonstrates that adding noise significantly enhances the robustness and desirability of the final solutions in the DTLZ7 and $\rho MNK$ problems. However, the impact of noise on DTLZ1 and DTLZ2 is less pronounced. This behavior can be explained by the bounded nature of their objective functions, which are confined to the interval $[0, 1]$. When these objectives converge to local optima, they tend to approach values near zero. Consequently, the magnitude of the noise introduced also becomes negligible, leading to minimal impact on the optimization process in these cases. This analysis underscores that the effect of noise is indeed significant in certain problem classes. As noise was fixed to a specific percentage of the objective values in this study, future research could explore the influence of varying noise levels on optimization performance. Investigating the effects of different magnitudes of noise could provide deeper insights into optimizing noise addition strategies, potentially improving solution robustness across a broader range of problems.   

\subsection{Investigation of Active and Relevant Objectives}
\begin{figure*}[h]
    \centering
    \includegraphics[width=\textwidth]{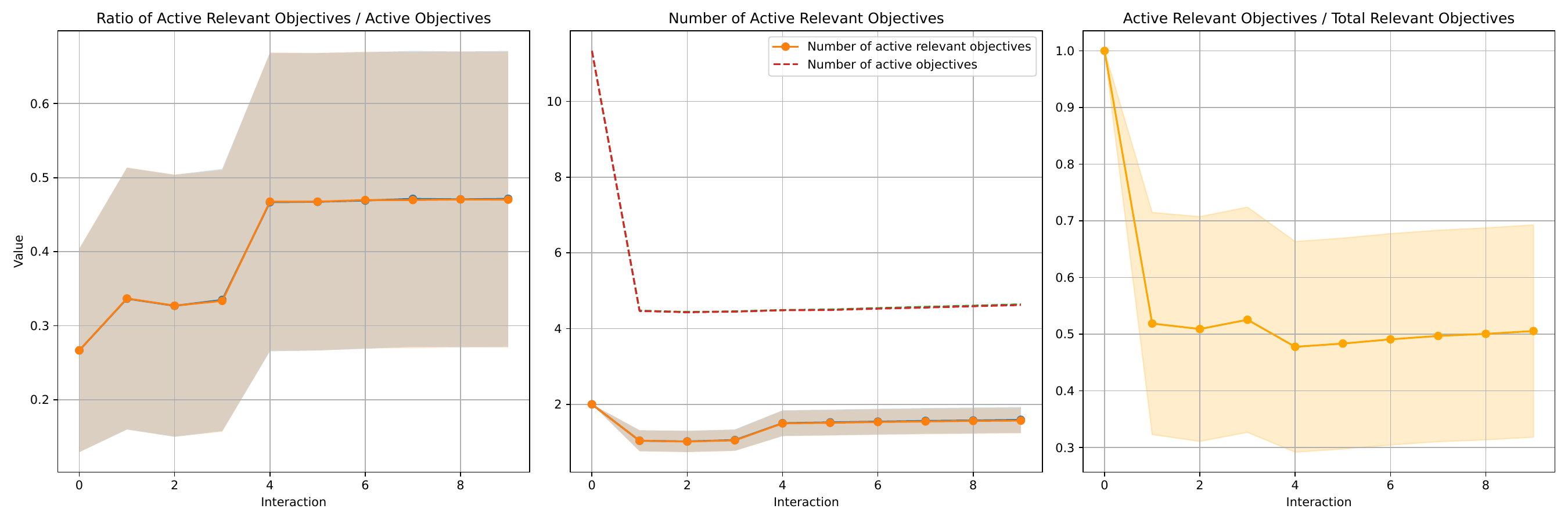}
    \caption{Evaluating the dynamics of active and relevant objectives across interactions. The shaded areas represent the standard deviation.}
    \label{fig:n_active}
\end{figure*}

The investigation of active and relevant objectives across interactions is crucial to understanding how the algorithm adapts and refines the set of objectives over time. This analysis is presented in Figure~\ref{fig:n_active}, which provides three graphs focusing on different metrics: the ratio of active relevant objectives to the total number of active objectives, the total number of active relevant objectives, and the ratio of active relevant objectives to the total number of relevant objectives.

Graph 1: The first graph shows how the ratio of active relevant objectives to the total number of active objectives increases over interactions. This indicates that the detection method effectively identifies and focuses on the most relevant objectives as the optimization progresses. This trend reflects improved efficiency, as the optimization process increasingly concentrates on objectives that matter most, leading to more relevant solutions and better resource utilization. It also highlights the system's ability to adapt based on user interactions, ultimately enhancing the overall quality.

Graph 2: The second graph demonstrates that while the total number of active objectives decreases, the number of active relevant objectives increases simultaneously. This trend highlights the method's effectiveness in refining the set of objectives to focus on those that truly matter to the DM. While the optimization starts with all the objectives being active, the algorithm eliminates irrelevant objectives after the first interaction with DM, resulting in more than $60\%$ savings in objective evaluations. By the fourth interaction, it has accurately identified the two relevant objectives, leaving a total of four active objectives. This is consistent with the first graph, where the ratio of active relevant objectives to total active objectives stabilizes around $0.5$, demonstrating the algorithm's effectiveness in focusing on the most important objectives.  

Graph 3: The third graph illustrates the ratio of active relevant objectives to total relevant objectives. Initially, this ratio is 1 in interaction 0, as all objectives are active. Over time, the algorithm selectively maintains focus on relevant objectives, showing a gradual and deliberate refinement process.

As observed in these graphs, the algorithm successfully identifies all relevant objectives ($\Fdm = 2$ ) starting from the fourth interaction. The shaded areas in all three graphs represent the standard deviation, providing insight into the variability of the results. The overall trend observed across these graphs supports the hypothesis that the algorithm effectively hones in on relevant objectives while reducing unnecessary complexity.

%%%%%%%%%%%%%%%%%%%%%%%%%
% To do:

% The effect detection, and reduction, in different drift change levels with Reset preferences (row problem with different objective numbers, column preference change levels

% Trend line showing the UF over interactions

% Checking for different UFs

% Number of objective evaluations in reduction methods

% Trend line for a number of active and number of relevant objectives in different interactions (in 4 lines for different thresholds when noise is on or off)

% Adding noise Effect

% Different problems

% Correlation between objectives in rMNK, also effect of K

%%%%%%%%%%%%%%%%%%%%%%%%%

\section{Conclusion and Future Work} \label{conclusion}
This study builds on previous research into hidden and irrelevant objectives within the context of iEMOAs. It advances our understanding by incorporating the dynamic nature of decision-maker (DM) preferences. To achieve this, a heuristic method was developed to simulate shifts in preferences across various intensities. Additionally, three distinct methods were examined to effectively manage these preference changes.

Based on these findings, several key insights have emerged from the analysis. First, the evaluation of various detection and reduction methods has shown that enabling the detection process significantly enhances the algorithm's performance, particularly when coupled with the reduction mechanism. As the reduction method activates all objectives from the start, there is an initial decline in utility values, but the algorithm rapidly compensates for this through iterative improvement. Ultimately, the integration of these methods leads to superior solutions, as demonstrated across different problem types. The results suggest that the dynamic nature of the reduction contributes to higher exploration and, consequently, more desirable outcomes, even surpassing the baseline scenario in later interactions. This underscores the robustness and adaptability of the proposed methodology.

The investigation into preference changes further highlights the advantages of employing dynamic reset mechanisms. The findings illustrate that the algorithm, when equipped with dynamic resets, outperforms other methods in scenarios involving severe preference shifts. The resilience of the dynamic method is evident in its ability to maintain high utility values, even as the intensity of preference changes increases. Compared to the fixed method, which discards historical preference data after each interaction, the dynamic reset demonstrates superior adaptability and efficiency in managing evolving decision-maker preferences. 

Regarding the addition of noise to fixed objective values, the results reveal that this technique enhances the algorithm’s ability to prevent misclassification of objectives in certain problem types. Specifically, noise addition proves effective in improving solution robustness in problems such as DTLZ7 and $\rho MNK$, where the introduction of noise contributes to better outcomes by mitigating the challenges posed by local optima. However, the limited impact of noise on problems like DTLZ1 and DTLZ2, where objective values approach zero, highlights the need for further exploration into optimizing noise levels for different problem contexts.

Finally, the investigation of active and relevant objectives reveals that the algorithm successfully refines its focus on the most pertinent objectives over successive interactions. The ratio of active relevant objectives to total active objectives increases over time, reflecting improved efficiency and the system's adaptability. This refinement is evident as the algorithm reduces the total number of active objectives while increasing the number of active relevant objectives, leading to substantial savings in computational resources. By the fourth interaction, the algorithm identifies the relevant objectives with high precision, demonstrating its effectiveness in isolating and addressing the decision-maker’s most important concerns.

Overall, the findings of this study suggest that incorporating dynamic preference behavior in iEMOAs results in more efficient and relevant optimization outcomes. We expect that these results will advance the development of dynamic methods that more effectively refine the set of active objectives and adapt to preference changes in evolving decision-making environments. To support further research, we will make our code publicly available at \texttt{www.to-be-updated-after-peer-review.com}.% (please note that due to the double-blind review process, the link will be published after the paper is accepted).
 %To support further research, we will make our code publicly available at \textcolor{red}{url to be added}.

This study was limited in several aspects that can be pursued in future studies. %Future research can expand this study in several key directions. 
First, varying noise levels in optimization performance should be explored further. While noise was fixed at a specific percentage in this study, testing different magnitudes could improve noise addition strategies, enhancing robustness across more problem types. Additionally, the DTLZ test suite has notable limitations, including its reliance on a single distance function, leading to dominance-resistant solutions \cite{ishibuchi2020effects}. In some cases, optimizing a subset of objectives improves the rest, oversimplifying the many-objective landscape \cite{brockhoff2007improving}. Future research should focus on more diverse and realistic test problems to overcome these issues.

While the dynamic reset method effectively handles preference drift, its performance could be improved by retaining preference data from previous interactions that align with the current preferences.

Although uni-variate feature selection can efficiently detect hidden and irrelevant objectives, the effects of incorporating nonlinear feature selection methods, such as multivariate or regression models, could be examined in future research. Finally, it is possible to implement a mechanism for early termination of the optimization to further reduce the number of objective evaluations. 
% Finally, future work might explore integrating alternative machine learning models within the iEMOA framework. Moving beyond SVM-based models to more general approaches could make the system more adaptable across different problem types, contributing to more robust and efficient optimization techniques.

%This research has introduced a dynamic approach to detecting hidden and irrelevant objectives within iEMOAs. By incorporating methods that account for the DM's evolving preferences, our approach enhances the adaptability and effectiveness of the optimization process. The integration of noise to mitigate the misclassification of fixed relevant objectives and the dynamic reset mechanism have been particularly effective, resulting in more robust and desirable solutions.

%Our experimental results demonstrate that the proposed method significantly reduces the number of objectives that are optimized, improves the utility of the solutions, and outperforms traditional methods by adapting to preference changes. This study's findings suggest that accounting for dynamic preference behavior in iEMOAs leads to more efficient and relevant optimization outcomes. We anticipate that the results of this study pave the way toward dynamic methods that are more effective in refining the set of active objectives and adapting to the preference drift that occurs in a dynamic decision-making environment. To this end, we will encourage further research by making our codes publicly available at \textcolor{red}{url to be added}.

%\bibliographystyle{IEEEtranN}
{\footnotesize
  \bibliographystyle{IEEEtranN}
  \bibliography{bib/abbrev,bib/journals,bib/authors,bib/articles.bib,bib/biblio,bib/my_bib.bib,bib/crossref}
}

\clearpage
% 
% \maketitle
% \include{Appendix.tex}
% that's all folks
\end{document}